# Improving the Efficiency of Inductive Logic Programming Through the Use of Query Packs

**Hendrik Blockeel**                                         HENDRIK.BLOCKEEL@CS.KULEUVEN.AC.BE
*Katholieke Universiteit Leuven, Department of Computer Science*
*Celestijnenlaan 200A, B-3001 Leuven, Belgium*

**Luc Dehaspe**                                              LUC.DEHASPE@PHARMADM.COM
*PharmaDM, Ambachtenlaan 54D, B-3001 Leuven, Belgium*

**Bart Demoen**                                              BART.DEMOEN@CS.KULEUVEN.AC.BE
**Gerda Janssens**                                           GERDA.JANSSENS@CS.KULEUVEN.AC.BE
**Jan Ramon**                                                JAN.RAMON@CS.KULEUVEN.AC.BE
*Katholieke Universiteit Leuven, Department of Computer Science*
*Celestijnenlaan 200A, B-3001 Leuven, Belgium*

**Henk Vandecasteele**                                       HENK.VANDECASTEELE@PHARMADM.COM
*PharmaDM, Ambachtenlaan 54D, B-3001 Leuven, Belgium*

## Abstract

Inductive logic programming, or relational learning, is a powerful paradigm for machine learning or data mining. However, in order for ILP to become practically useful, the efficiency of ILP systems must improve substantially. To this end, the notion of a query pack is introduced: it structures sets of similar queries. Furthermore, a mechanism is described for executing such query packs. A complexity analysis shows that considerable efficiency improvements can be achieved through the use of this query pack execution mechanism. This claim is supported by empirical results obtained by incorporating support for query pack execution in two existing learning systems.

## 1. Introduction

Many data mining algorithms employ to some extent a generate-and-test approach: large amounts of partial or complete hypotheses are generated and evaluated during the data mining process. This evaluation usually involves testing the hypothesis on a large data set, a process which is typically linear in the size of the data set. Examples of such data mining algorithms are APRIORI (Agrawal et al., 1996), decision tree algorithms (Quinlan, 1993a; Breiman et al., 1984), algorithms inducing decision rules (Clark & Niblett, 1989), etc.

Even though the search through the hypothesis space is seldom exhaustive in practical situations, and clever branch-and-bound or greedy search strategies are employed, the number of hypotheses generated and evaluated by these approaches may still be huge. This is especially true when a complex hypothesis space is used, as is often the case in inductive logic programming (ILP), where the sheer size of the hypothesis space is an important contribution to the high computational complexity of most ILP approaches. This computational complexity can be reduced, however, by exploiting the fact that there are many similarities between hypotheses.

Most ILP systems build a hypothesis one clause at a time. This search for a single clause is what we will be concerned with in the rest of this paper, and so the word "hypothesis" further on will usually refer to a single clause. The clause search space is typically structured as a lattice. Because clauses close to one another in the lattice are similar, the computations involved in evaluating them will be similar as well. In other words, many of the computations that are performed when evaluating one clause (which boils down to executing a query consisting of the body of the clause) will have to be performed again when evaluating the next clause. Storing certain intermediate results during the computation for later use could be a solution (e.g., tabling as in the XSB Prolog engine, Chen & Warren, 1996), but may be infeasible in practice because of its memory requirements. It becomes more feasible if the search is reorganised so that intermediate results are always used shortly after they have been computed; this can be achieved to some extent by rearranging the computations. The best way of removing the redundancy, however, seems to be to re-implement the execution strategy of the queries in such a way that as much computation as possible is effectively shared.

In this paper we discuss a strategy for executing sets of queries, organised in so-called query packs, that avoids the redundant computations. The strategy is presented as an adaptation of the standard Prolog execution mechanism. The adapted execution mechanism has been implemented in ilProlog, a Prolog system dedicated to inductive logic programming. Several inductive logic programming systems have been re-implemented to make use of this dedicated engine, and using these new implementations we obtained experimental results showing in some cases a speed-up of more than an order of magnitude. Thus, our work significantly contributes to the applicability of inductive logic programming to real world data mining tasks. In addition, we believe it may contribute to the state of the art in query optimisation in relational databases. Indeed, in the latter field there has been a lot of work on the optimisation of individual queries or relatively small sets of queries, but much less on the optimisation of large groups of very similar queries, which understandably did not get much attention before the advent of data mining. Optimisation of groups of queries for relational databases seems an interesting research area now, and we believe techniques similar to the ones proposed here might be relevant in that area.

The remainder of this paper is structured as follows. In Section 2 we precisely describe the ILP problem setting in which this work is set. In Section 3 we define the notion of a query pack and indicate how it would be executed by a standard Prolog interpreter and what computational redundancy this causes. We further describe an execution mechanism for query packs that makes it possible to avoid the redundant computations that would arise if all queries in the pack were run separately, and show how it can be implemented by making a few small but significant extensions to the WAM, the standard Prolog execution mechanism. In Section 4 we describe how the query pack execution strategy can be incorporated in two existing inductive logic programming algorithms (Tilde and Warmr). In Section 5 we present experimental results that illustrate the speed-up that these systems achieve by using the query pack execution mechanism. In Section 6 we discuss related work and in Section 7 we present conclusions and some directions for future work.

## 2. Inductive Logic Programming

Inductive logic programming (Muggleton & De Raedt, 1994) is situated in the intersection of machine learning or data mining on the one hand, and logic programming on the other hand. It shares with the former fields the goal of finding patterns in data, patterns that can be used to build predictive models or to gain insight in the data. With logic programming it shares the use of clausal first order logic as a representation language for both data and hypotheses. In the remainder of this text we will use some basic notions from logic programming, such as literals, conjunctive queries, and variable substitutions. We will use Prolog notation throughout the paper. For an introduction to Prolog and logic programming see Bratko (1990).

Inductive logic programming can be used for many different purposes, and the problem statements found in ILP papers consequently vary. In this article we consider the so-called learning from interpretations setting (De Raedt & Džeroski, 1994; De Raedt, 1997). It has been argued elsewhere that this setting, while slightly less powerful than the standard ILP setting (it has problems with, e.g., learning recursive predicates), is sufficient for most practical purposes and scales up better (Blockeel et al., 1999).

We formulate the learning task in such a way that it covers a number of different problem statements. More specifically, we consider the problem of detecting for a set of conjunctive queries for which instantiations of certain variables each query succeeds. These variables are called *key variables*, and a grounding substitution for them is called a *key instantiation*. The intuition is that an example in the learning task is uniquely identified by a single key instantiation.

The link with ILP systems that learn clauses is then as follows. The search performed by an ILP system is directed by regularly evaluating candidate clauses. Let us denote such a candidate clause by $Head(X) \leftarrow Body(X, Y)$ where $X$ represents a vector of variables appearing in the head of the clause and $Y$ represents additional variables that occur in the body. We assume that the head is a single literal and that a list of examples is given, where each example is of the form $Head(X)\theta$ with $\theta$ a substitution that grounds $X$. Examples may be labelled (e.g., as positive or negative), but this is not essential in our setting. While an example can be represented as a fact $Head(X)\theta$ when learning definite Horn clauses, we can also consider it just a tuple $X\theta$. Both notations will be used in this paper.

Intuitively, when positive and negative examples are given, one wants to find a clause that covers as many positive examples as possible, while covering few or no negatives. Whether a single example $Head(X)\theta$ is covered by the clause or not can be determined by running the query $? - Body(X, Y)\theta$. In other words, evaluating a clause boils down to running a number of queries consisting of the body of the clause. For simplicity of notation, we will often denote a conjunctive query by just the conjunction (without the $?-$ symbol).

In some less typical ILP settings, the ILP algorithm does not search for Horn clauses but rather for general clauses, e.g., CLAUDIEN (De Raedt & Dehaspe, 1997) or for frequent patterns that can be expressed as conjunctive queries, e.g., WARMR(Dehaspe & Toivonen, 1999). These settings can be handled by our approach as well: all that is needed is a mapping from hypotheses to queries that allow to evaluate these hypotheses. Such a mapping is defined by De Raedt and Dehaspe (1997) for CLAUDIEN; for WARMR it is trivial.

Given a set of queries $S$ and a set of examples $E$, the main task is to determine which queries $Q \in S$ cover which examples $e \in E$. We formalise this using the notion of a result set:

**Definition 1 (Result set)** *The* result set *of a set of queries $S$ in a deductive database $D$ for key $K$ and example set $E$, is*

$$RS(S, K, D, E) = \{(K\theta, i) | Q_i \in S \text{ and } K\theta \in E \text{ and } Q_i\theta \text{ succeeds in } D\}$$

Similar to the learning from interpretations setting defined in (De Raedt, 1997), the problem setting can now be stated as:

**Given:** a set of conjunctive queries $S$, a deductive database $D$, a tuple $K$ of variables that occur in each query in S, and an example set $E$

**Find:** the result set $RS(S, K, D, E)$; i.e., find for each query $Q$ in $S$ those ground instantiations $\theta$ of $K$ for which $K\theta \in E$ and $Q\theta$ succeeds in $D$.

**Example 1** *Assume an ILP system learning a definition for* grandfather/2 *wants to evaluate the following hypotheses:*

```
grandfather(X,Y) :- parent(X,Z), parent(Z,Y), male(X).
grandfather(X,Y) :- parent(X,Z), parent(Z,Y), female(X).
```

*Examples are of the form* grandfather(*gf*,*gc*) *where gf and gc are constants; hence each example is uniquely identified by a ground substitution of the tuple $(X, Y)$. So in the above problem setting the set of Prolog queries $S$ equals* {(?- parent(X,Z), parent(Z,Y), male(X)), (?- parent(X,Z), parent(Z,Y), female(X))} *and the key $K$ equals $(X, Y)$. Given a query $Q_i \in S$, finding all tuples $(x, y)$ for which $((x, y), i) \in R$ (with $R$ the result set as defined above) is equivalent to finding which of the* grandfather(*x*,*y*) *facts in the example set are predicted by the clause* grandfather(X,Y) :- $Q_i$.

The generality of our problem setting follows from the fact that once it is known which queries succeed for which examples, the statistics and heuristics that typical ILP systems use can be readily obtained from this. A few examples:

- discovery of frequent patterns (Dehaspe & Toivonen, 1999): for each query $Q_i$ the number of key instantiations for which it succeeds just needs to be counted, i.e., $freq(Q_i) = |\{K\theta | (K\theta, i) \in R\}|$ with $R$ the result set.

- induction of Horn clauses (Muggleton, 1995; Quinlan, 1993b): the accuracy of a clause $H$ :- $Q_i$ (defined as the number of examples for which body and head hold, divided by the number of examples for which the body holds) can be computed as $\frac{|\{K\theta | (K\theta, i) \in R \wedge D \models H\theta\}|}{|\{K\theta | (K\theta, i) \in R\}|}$ with $R$ the result set.

- induction of first order classification or regression trees (Kramer, 1996; Blockeel & De Raedt, 1998; Blockeel et al., 1998): the class entropy or variance of the examples covered (or not covered) by a query can be computed from the probability distribution of the target variable; computing this distribution involves simple counts similar to the ones above.

After transforming the *grandfather/2* clauses into

```
grandfather((X,Y)),I) :- parent(X,Z), parent(Z,Y), male(X), I = 1.
grandfather((X,Y)),I) :- parent(X,Z), parent(Z,Y), female(X), I = 2.
```

the result set can clearly be computed by collecting for all grounding $\theta$'s where $K\theta \in E$ the answers to the query ?- `grandfather(`$K\theta$`,I)` . In Section 3 the queries will have a literal `I = i` at the end or another goal which by side-effects results in collecting the result set.

In practice, it is natural to compute the result set using a double loop: one over examples and one over queries and one has the choice as to which is the outer loop. Both the "examples in outer loop" and the "queries in outer loop" have been used in data mining systems; in the context of decision trees, see for instance Quinlan (1993a) and Mehta et al. (1996). We shall see further that the redundancy removal approach we propose uses the "examples in outer loop" strategy. In both approaches however, given a query and a key instantiation, we are interested only in whether the query succeeds for that key instantiation. This implies that after a particular query has succeeded on an example, its execution can be stopped.

In other words: computing the result set defined above boils down to evaluating each query on each example, where we are only interested in the existence of success for each such evaluation. Computing more than one solution for one query on one example is unnecessary.

## 3. Query Packs

For simplicity, we make abstraction of the existence of keys in the following examples. What is relevant here, is that for each query we are only interested in whether it succeeds or not, not in finding all answer substitutions.

Given the following set of queries

```
p(X), I = 1.
p(X), q(X,a), I = 2.
p(X), q(X,b), I = 3.
p(X), q(X,Y), t(X), I = 4.
p(X), q(X,Y), t(X), r(Y,1), I = 5.
```

we can choose to evaluate them separately. Since we are only interested in one – the first – success for each query, we would evaluate in Prolog the queries

```
once((p(X), I = 1)).
once((p(X), q(X,a), I = 2)).
once((p(X), q(X,b), I = 3)).
once((p(X), q(X,Y), t(X), I = 4)).
once((p(X), q(X,Y), t(X), r(Y,1), I = 5)).
```

The wrapper *once/1* is a pruning primitive and prevents the unnecessary search for more solutions. Its definition in Prolog is simply

```
once(Goal) :- call(Goal), !.
```

An alternative way to evaluate the queries consists in merging them into one (nested) disjunction as in:

```
p(X), (I=1 ; q(X,a), I=2 ; q(X,b), I=3 ; q(X,Y), t(X), (I=4 ; r(Y,1), I=5)).
```

The set of queries can now be evaluated as a whole: the success of one branch in the disjunctive query corresponds to the success of the corresponding individual query.

Compared to the evaluation of the individual queries, the disjunctive query has both an advantage and a disadvantage:

+ all the queries have the same prefix p(X), which is evaluated once in **each** individual query, while in the disjunctive query, the goal p(X) is evaluated only once; depending on the evaluation cost of p/1, this can lead to arbitrary performance gains.

− the usual Prolog pruning primitives are not powerful enough to prevent all the unnecessary backtracking after a branch in the disjunctive query has succeeded; this is explained further in Example 2.

**Example 2** *In this example the literals $I = i$ have been left out, because they do not contribute to the discussion:*

```
p(X), q(X).
p(X), r(X).
```

*Evaluating these queries separately means evaluating*

```
once((p(X), q(X))).
once((p(X), r(X))).
```

*or equivalently*

```
p(X), q(X), !.
p(X), r(X), !.
```

*The corresponding disjunctive query is*

```
p(X), (q(X) ; r(X)).
```

*We can now try to place a pruning primitive in the disjunctive query: !/0 at the end of each branch results in*

```
p(X), (q(X), ! ; r(X), !)
```

*The scope of the first cut is clearly too large: after the goal q(X) has succeeded, the cut will prevent entering the second branch. It means that adding the cut in the disjunctive query leads to a wrong result.*

*Using* once/1 *in the disjunctive query results in*

```
p(X), (once(q(X)) ; once(r(X)))
```

*This results in a correct query. However, both branches are still executed for every binding that the goal* `p(X)` *produces, even if both branches have succeeded already.*

The combination of the advantage of the disjunctive query with the advantage of the individual query with pruning (once or cut) results in the notion of the *query pack*. Syntactically, a query pack looks like a disjunctive query where the ; control construct is replaced by a new control construct denoted by *or*. So the query pack corresponding to the disjunctive query above is

`p(X), (I=1 or q(X,a), I=2 or q(X,b), I=3 or q(X,Y), t(X), (I=4 or r(Y,1), I=5))`

This query pack can be represented as the tree in Figure 1. For a query pack $\mathcal{Q}$ such a tree has literals or conjunctions of literals in the nodes. Each path from the root to a leaf node represents a conjunctive query $Q$ which is a *member* of $\mathcal{Q}$, denoted $Q \in \mathcal{Q}$. The *or* construct is implicit in the branching points.

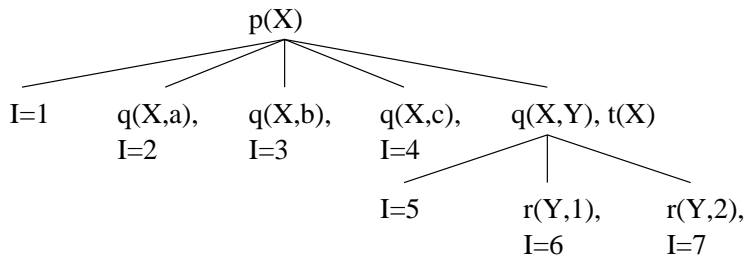

Figure 1: A query pack.

The intended procedural behaviour of the *or* construct is that once a branch has succeeded, it is effectively pruned away from the pack during the evaluation of the query pack on the current example. This pruning must be recursive, i.e., when all branches in a subtree of the query pack have succeeded, the whole subtree must be pruned. Evaluation of the query pack then terminates when all subtrees have been pruned or all of the remaining queries fail for the example.

The semantics of the *or* construct and its efficient implementation is the subject of the rest of this section. It should however be clear already now that in the case that all the answers of each query are needed, pruning cannot be performed and the disjunctive query is already sufficient, i.e., query packs are useful when a single success per query suffices.

### 3.1 Efficient Execution of Query Packs

In Section 3.1.2, a meta-interpreter is given that defines the behaviour of query packs. In practice this meta-interpreter is not useful, because in many cases the meta-interpreter itself causes more overhead than the use of query packs can compensate for. Indeed, previously reported results (Demoen et al., 1999; Blockeel, 1998) indicate that the overhead involved in a high-level Prolog implementation destroys the efficiency gain obtained by redundancy reduction. Moreover as discussed in Section 3.1.2, the meta-interpreter does not have the desired time-complexity. This shows that the desired procedural semantics of *or* can be implemented in Prolog itself, but not with the desired performance because Prolog lacks the appropriate primitives.

The conclusion is that changes are needed at the level of the Prolog engine itself. This requires an extension of the WAM (Warren Abstract Machine) which is the underlying abstract machine for most Prolog implementations. The extended WAM provides the *or* operator as discussed above: it permanently removes branches from the pack that do not need to be investigated anymore. This extended WAM has become the basis of a new Prolog engine dedicated to inductive logic programming, called ILPROLOG. This section continues with the introduction of some basic terminology for query packs and explains at a high level how query pack execution works. Next our meta-interpreter for the query pack execution is given and finally the changes needed for the WAM are clarified.

### 3.1.1 Principles of Query Packs (Execution)

Before we discuss query pack execution in detail, note the following two points: (1) during the pack execution, the pruning of a branch must survive backtracking; (2) when executing a pack we are not interested in any variable instantiations, just in whether a member of the pack succeeds or not. In our previous description we were interested in the binding to the variable I. Since each branch can bind I to only one value – the query number – we collect these values in practice by a side effect denoted in Section 3.2 by *report_success*.

The starting point for the query pack execution mechanism is the usual Prolog execution of a query $Q$ given a Prolog program $P$. By backtracking Prolog will generate all the solutions for $Q$ by giving the possible instantiations $\theta$ such that $Q\theta$ succeeds in $P$.

A query pack consists of a conjunction of literals and a set of *alternatives*, where each alternative is again a query pack. Note that leaves are query packs with an empty set of alternatives. For each query pack $\mathcal{Q}$, $conj(\mathcal{Q})$ denotes the conjunction and $children(\mathcal{Q})$ denotes the set of alternatives. A set of queries is then represented by a so-called *root* query pack. For every query pack $\mathcal{Q}$, there is a path of query packs starting from the root query pack $\mathcal{Q}_{root}$ and ending at the query pack itself, namely $< \mathcal{Q}_{root}, \mathcal{Q}_1, ..., \mathcal{Q}_n, \mathcal{Q} >$. The query packs in this path are the *predecessors* of $\mathcal{Q}$. Every query pack has a set of *dependent* queries, $dependent\_queries(\mathcal{Q})$. Let $< \mathcal{Q}_{root}, \mathcal{Q}_{i_1}, ..., \mathcal{Q}_{i_n}, \mathcal{Q} >$ be the path to $\mathcal{Q}$, then $dependent\_queries(\mathcal{Q}) = \{conj(\mathcal{Q}_{root}) \wedge conj(\mathcal{Q}_{i_1}) \wedge ... \wedge conj(\mathcal{Q}_{i_n}) \wedge conj(\mathcal{Q}) \wedge conj(\mathcal{Q}_{j_1}) \wedge ... \wedge conj(\mathcal{Q}_{j_m}) \wedge conj(\mathcal{Q}_l) \mid < \mathcal{Q}, \mathcal{Q}_{j_1}, ..., \mathcal{Q}_{j_m}, \mathcal{Q}_l >$ is a path from $\mathcal{Q}$ to a leaf $\mathcal{Q}_l\}$. Note that $dependent\_queries(\mathcal{Q}_{root})$ are actually the members of the query pack as described earlier.

**Example 3** *For the query pack in Figure 1, $\mathcal{Q}_{root}$ is the root of the tree. $conj(\mathcal{Q}_{root})$ is $p(X)$. The set $children(\mathcal{Q}_{root})$ contains the 4 query packs which correspond to the trees rooted at the 4 sons of the root of the tree. Suppose that these query packs are named (from left to right) $\mathcal{Q}_1, \mathcal{Q}_2, \mathcal{Q}_3,$ and $\mathcal{Q}_4$. Then $conj(\mathcal{Q}_2)$ equals $(q(X,a), I = 2)$, $children(\mathcal{Q}_2)$ equals the empty set, $conj(\mathcal{Q}_4)$ equals $(q(X,Y), t(X))$, and $dependent\_queries(\mathcal{Q}_4)$ equals $\{(p(X), q(X,Y), t(X), I = 4), (p(X), q(X,Y), t(X), r(Y,1), I = 5)\}$.*

Execution of a root query pack $\mathcal{Q}_{root}$ aims at finding out which queries of the set $dependent\_queries(\mathcal{Q}_{root})$ succeed. If a query pack is executed as if the *or*s were usual disjunctions, backtracking occurs over queries that have already succeeded and too many

```
0    execute_qp( pack Q, substitution θ) {
1      while ( σ ← next_solution( conj(Q)θ)
2         {
3            for each Q_child in children(Q) do
4               {
5                  if ( execute_qp( Q_child , σ) == success)
6                     children(Q) ← children(Q) \ {Q_child}
7               }
8            if ( children(Q) is an empty set) return(success)
9         }
10     return(fail)
11   }
```

Figure 2: The query pack execution algorithm.

successes are detected. To avoid this, it should be the case that as soon as a query succeeds, the corresponding part of the query pack should no longer be considered during backtracking. Our approach realises this by reporting success of queries (and of query packs) to predecessors in the query pack. A (non-root) query pack $Q$ can be safely removed if all the queries that depend on it (i.e., all the queries in $dependent\_queries(Q)$) succeeded once. For a leaf $Q$ (empty set of children), success of $conj(Q)$ is sufficient to remove it. For a non-leaf $Q$, we wait until all the dependent queries report success or equivalently until all the query packs in $children(Q)$ report success.

At the start of the evaluation of a root query pack, the set of children for every query pack in it contains all the alternatives in the given query pack. During the execution, query packs can be removed from children sets and thus the values of the $children(Q)$ change accordingly. When due to backtracking a query pack is executed again, it might be the case that fewer alternatives have to be considered.

The execution of a query pack $Q\theta$ is defined by the algorithm $execute\_qp(Q, \theta)$ (Figure 2) which imposes additional control on the usual Prolog execution.

The usual Prolog execution and backtracking behaviour is modelled by the while loop (line 1) which generates all possible solutions $\sigma$ for the conjunction in the query pack. If no more solutions are found, fail is returned and backtracking will occur at the level of the calling query pack.

The additional control manages the $children(Q)$. For each solution $\sigma$, the necessary children of $Q$ will be executed. It is important to notice that the initial set of children of a query pack is changed destructively during the execution of this algorithm. Firstly, when a leaf is reached, success is returned (line 8) and the corresponding child is removed from the query pack (line 6). Secondly, when a query pack that initially had several children, finally ends up with an empty set of children (line 6), also this query pack is removed (line 8). The fact that children are destructively removed, implies that when due to backtracking the same query pack is executed again for a different $\sigma$, not all of the alternatives that were initially there, have to be executed any more. Moreover, by returning success the backtracking over the current query pack conjunction $conj(\mathcal{Q})$ is stopped: all branches have reported success.

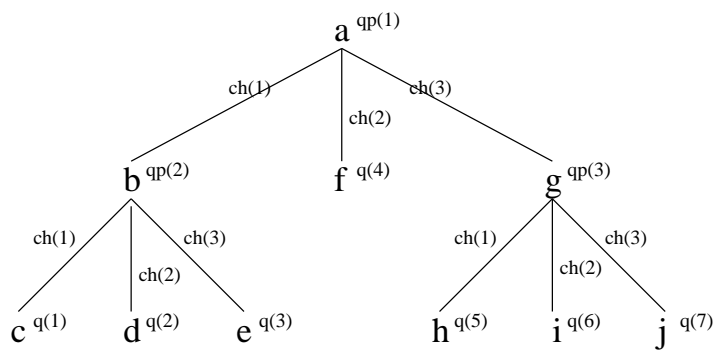

Figure 3: Query pack numbers $qp(i)$, Query numbers $q(i)$ and Child numbers $ch(i)$ in our example.

### 3.1.2 A Meta-interpreter for Query Packs

The first implementation of the query pack execution algorithm is the meta-interpreter $meta\_execute\_qp(Q)$. The meta-interpreter uses the following labelling in its representation of a query pack:

- **Query pack number** All the non-leaf query packs in the tree are numbered, depth first, from left to right $(qp(i))$.

- **Query number** Each leaf is numbered, from left to right. If the original queries were numbered sequentially, then the numbers at the leaves correspond with these $(q(i))$.

- **Child number** For each non-leaf query pack with N children, all children are numbered from 1 up to N sequentially $(ch(i))$.

Consider the query pack $a$, $(b, (c\ or\ d\ or\ e)\ or\ f\ or\ g, (h\ or\ i\ or\ j))$. Note that the atoms in the example could in general be arbitrary conjunctions of non-ground terms. Its labelling is shown in Figure 3.

A labelled query pack $\mathcal{Q}$ is then represented as a Prolog term as follows (with $\mathcal{Q}_f$ the father of $\mathcal{Q}$):

- A leaf $\mathcal{Q}$ is represented by the term $(c, leaf(qpnbf, chnb, qnb))$ with $c$ the $conj(\mathcal{Q})$, $qpnbf$ the query pack number of $\mathcal{Q}_f$, $chnb$ the child number of $\mathcal{Q}$ w.r.t. $\mathcal{Q}_f$, and $qnb$ the query number of $\mathcal{Q}$.

- A non-leaf $\mathcal{Q}$ is represented by the term $(c, or(cs, qpnbf, qpnb, chnb, totcs)$ with $c$ the $conj(\mathcal{Q})$, $cs$ the list $children(\mathcal{Q})$, $qpnbf$ the query pack number of $\mathcal{Q}_f$, $qpnb$ the query pack number of $\mathcal{Q}$, $chnb$ the child number of $\mathcal{Q}$ w.r.t. $\mathcal{Q}_f$, and $totcs$ the total number of $children(\mathcal{Q}))$. The query pack number of the father of the root query pack is assumed to be zero.

The example of Figure 3 has the following representation (as a Prolog term):

```
(a, or([(b,or([(c,leaf(2,1,1)),(d,leaf(2,2,2)),(e,leaf(2,3,3))],1,2,1,3)),
       (f,leaf(1,2,4)),
       (g,or([(h,leaf(3,1,5)),(i,leaf(3,2,6)),(j,leaf(3,3,7))],1,3,3,3))],
   0,1,1,3))
```

During the execution of the meta-interpreter, solved/2 facts are asserted. Each fact solved(*qpnb*, *chnb*) denotes that the child with number *chnb* from query pack with number *qpnb* has succeeded. Such facts are asserted when reaching a leaf and also when all children of a query pack have succeeded. The meta-interpreter only executes children for which no solved/2 fact has been asserted.

Note that the time-complexity of this meta-interpreter is not yet as desired. Execution of a query pack will always be dependent on the number of original children, instead of the number of remaining (as yet unsuccessful) children.

```
run_QueryPack(Q) :-
      preprocess(Q, Qlabeled, 0, 1, 1, 1, _, _),
             % The code for preprocessing is given in Appendix A
      retractall(solved(_, _)),
      meta_execute_qp(Qlabeled),
      solved(0, _), !.

meta_execute_qp((A,B)) :- !,
      call(A),
      meta_execute_qp(B).
meta_execute_qp(or(Cs, QpNbF, QpNb, ChildNb, TotCs)) :-
      !, % 'or' corresponds to a non-leaf query pack
      handlechildren(Cs, QpNb, 1),
      all_solved(QpNb, 0, TotCs),
      assert(solved(QpNbF,ChildNb)).
meta_execute_qp(leaf(QpNbF, ChildNb , QueryNb)) :-
      !, % 'leaf' corresponds to the end of a query
      write(succeed(QueryNb)), nl,
      assert(solved(QpNbF,ChildNb)).

handlechildren([], _, _).
handlechildren([C|_], QpNb, ChildNb) :-
      not(solved(QpNb,ChildNb)),
      once(meta_execute_qp(C)), fail.
handlechildren([_|Cs], QpNb, ChildNb) :-
      ChildNb1 is ChildNb + 1,
      handlechildren(Cs, QpNb, ChildNb1).

all_solved(QpNb, ChildNb, TotCs) :-
      (ChildNb = TotCs -> true
      ;    ChildNb1 is ChildNb + 1,
           solved(QpNb, ChildNb1),
           all_solved(QpNb, ChildNb1, TotCs)
      ).
```

### 3.1.3 WAM Extensions

To fully exploit the potential of a query pack (shared computation and avoidance of unnecessary backtracking) changes have to be made at the level of the Prolog engine itself. The explanation assumes a WAM-based Prolog engine (Aït-Kaci, 1991) but a short explanation of the execution of disjunction in Prolog is given first, so that it becomes more easy to see what was newly introduced in the WAM.

Assume that the body of a clause to be executed is *a, (b,c ; d ; e)*. Assume also that all predicates have several clauses. At the moment that execution has reached the first clause of *c*, the choice point stack looks like Figure 4(a): there are choice points for the activation of *a*, the disjunction itself, *b* and *c*. The choice points are linked together so that backtracking can easily pop the top most one. Each choice point contains a pointer to the next alternative to be tried: only for the disjunction choice point, this alternative pointer is shown. It points to the beginning of the second branch of the disjunction. After all alternatives for *b* and *c* have been exhausted, this second branch is entered and *d* becomes active: this is the situation shown in Figure 4(b). At that point, the alternative of the disjunction choice point refers to the last alternative branch of the disjunction. Finally, once *e* is entered, the disjunction choice point is already popped.

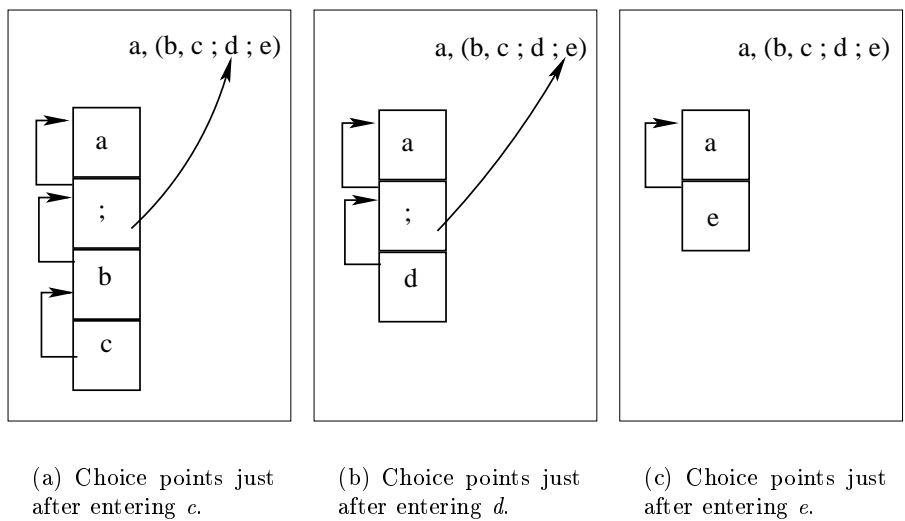

(a) Choice points just after entering *c*.  (b) Choice points just after entering *d*.  (c) Choice points just after entering *e*.

Figure 4: Illustration of execution of disjunction in the WAM.

When the goal *a* produces a new solution, all branches of the disjunction must be tried again. It is exactly this we want to avoid for query packs: a branch that has succeeded once, should never be re-entered. We therefore adapt the disjunction choice point to become an *or*-choice point which is set up to point into a data structure that contains references to each alternative in the *or* disjunction. This data structure is named the *pack table*. Figure 5(a) shows the state of the execution when it has reached *c*: it is similar to Figure 4(a). The *or*-choice point now contains the information that the first branch is being executed. As the execution proceeds, there are two possibilities: either this first branch succeeds or it fails. We describe the failing situation for the first branch and explain what happens on success of the second branch. If the first branch has no solution, backtracking updates the alternative in the *or*-choice point, to point to the next branch in the pack table. The situation after the second branch is entered is shown in 5(b) and is again similar to 4(b). Suppose now that the branch with the goal *d* succeeds: the entry in the pack table with *or*-alternatives is now adapted by erasing the second alternative branch, backtracking occurs, and the next alternative branch of the *or*-choice point is taken. This is shown in 5(c).

When *a* produces a new solution and the *or*-disjunction is entered again, the pack table does no longer contain the second alternative branch and it is never re-entered. The pack table is actually arranged in such a way that entries are really removed instead of just erased so that they cause no overhead later.

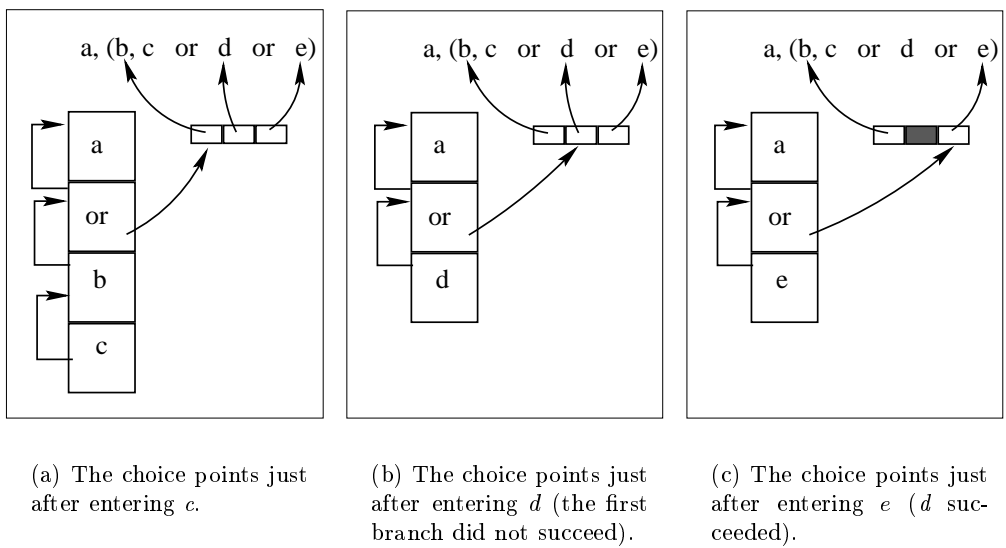

(a) The choice points just after entering *c*.

(b) The choice points just after entering *d* (the first branch did not succeed).

(c) The choice points just after entering *e* (*d* succeeded).

Figure 5: Illustration of execution of pack disjunction on the WAM.

Two more issues must be explained: first, the pack table with alternatives must be constructed at runtime every time the query pack is entered for evaluation. This is done by emitting the necessary instructions in the beginning of the code for the query pack. As an example, we show the code for the query pack *a, (b,c or d or e)* in Figure 6.

Finally, in the example it is clear that at the moment that all alternatives of an *or*-disjunction have succeeded, *a* can stop producing more solutions. So the computation can be stopped. In general - with nested query packs - it means that one pack table entry of the next higher *or*-node can be erased and this in a recursive way. The recursive removal of entries from the pack tables, is done by the instruction *query_pack_prune*.

We have implemented this schema in ILPROLOG. Section 5 presents some measurements in ILPROLOG.

### 3.2 Using Query Packs

Figure 7 shows an algorithm that makes use of the pack execution mechanism to compute the result set *R* as defined in our problem statement. The set *S* of queries is here typically

```
        construct_pack_table @1, @2, @3
        call a
        query_pack_try
   @1:  call b
        call c
        query_pack_prune
   @2:  call d
        query_pack_prune
   @3:  call e
        query_pack_prune
```

Figure 6: Abstract machine code for *a, (b,c or d or e)* .

the set of all refinements of a given query, i.e., it does not correspond to the whole hypothesis space. From a query pack $\mathcal{Q}$ containing all queries in $S$, a derived pack $\mathcal{Q}'$ is constructed by adding a report_success/2 literal to each leaf of the pack; the (procedural) task of report_success(K,$i$) is simply to add ($K$,$i$) to the result set $R$. Obviously a specific ILP system not interested in the result set itself could provide its own report_success/2 predicate and thus avoid the overhead of explicitly building the result set.[1]

```
1    evaluate(set of examples E, pack Q, key K) {
2        Q' ← Q;
3        q ← 1;
4        for each leaf of Q' do {
5            add report_success(K, q) to the right of the conjunction in the leaf
6            increment q
7        }
8        C ← (evaluate_pack(K) :- Q');
9        compile_and_load(C);
10       for each example e in E do {
11           evaluate_pack(e);
12       }
13   }
```

Figure 7: Using query packs to compute the result set.

Note that the algorithm in Figure 7 follows the strategy of running all queries for each single example before moving on to the next example: this could be called the "examples in outer loop" strategy, as opposed to the "queries in outer loop" strategy used by most ILP systems. The "examples in outer loop" strategy has important advantages when processing large data sets, mainly due to the ability to process them efficiently without having all data in main memory at the same time (Mehta et al., 1996; Blockeel et al., 1999).

---

1. In our current implementation the result set is implemented as a bit-matrix indexed on queries and examples. This implementation is practically feasible (on typical computers at the time of writing) even when the number of queries in the pack multiplied by the number of examples is up to a billion, a bound which holds for most current ILP applications.

### 3.3 Computational Complexity

We estimate the speedup factor that can be achieved using query pack execution in two steps: first we consider one-level packs, then we extend the results towards deeper packs.

Lower and upper bounds on the speedup factor that can be achieved by executing a one-level pack instead of separate queries can be obtained as follows. For a pack containing $n$ queries $q_i = (a, b_i)$, let $T_i$ be the time needed to compute the first answer substitution of $q_i$ if there are any, or to obtain failure otherwise. Let $t_i$ be the part of $T_i$ spent within $a$ and $t'_i$ the part of $T_i$ spent in $b_i$. Then $T_s = \sum_i (t_i + t'_i)$ and $T_p = \max(t_i) + \sum_i t'_i$ with $T_s$ representing the total time needed for executing all queries separately and $T_p$ the total time needed for executing the pack. Introducing $c = \sum_i t_i / \sum_i t'_i$, which roughly represents the ratio of the computational complexity in the shared part over that in the non-shared part, we have

$$\frac{T_s}{T_p} = \frac{\sum_i t_i + \sum_i t'_i}{\max_i t_i + \sum_i t'_i} = \frac{c+1}{\frac{\max_i t_i}{\sum_i t'_i} + 1} \tag{1}$$

Now defining $K$ as the ratio of the maximal $t_i$ over the average $t_i$, i.e.

$$K = \frac{\max_i t_i}{\sum_i t_i / n}$$

we can rewrite Equation (1) as

$$\frac{T_s}{T_p} = \frac{c+1}{\frac{K}{n}c + 1} \tag{2}$$

Since $\frac{\sum_i t_i}{n} \le \max t_i \le \sum_i t_i$ we know $1 \le K \le n$, which leads to the following bounds:

$$1 \le \frac{T_s}{T_p} \le \frac{c+1}{\frac{c}{n}+1} < \min(c+1, n) \tag{3}$$

Thus the speedup factor is bounded from above by the branching factor $n$ and by the ratio $c$ of computational complexity in the shared part over the computational complexity of the non-shared part; and a maximal speedup can be attained when $\max t_i \simeq \sum t_i / n$ (or, $K \simeq 1$), in other words when the $t_i$ for all queries are approximately equal.

For multi-level packs, we can estimate the efficiency gain as follows. Given a query $q_i$, let $T_i$ be defined as above (the total time for finding 1 answer to $q_i$ or obtaining failure). Instead of $t_i$ and $t'_i$, we now define $t_{i,l}$ as the time spent on level $l$ of the pack when solving $q_i$; counting the root as level 0 and denoting the depth of the pack with $d$ we have $T_i = \sum_{l=0}^{d} t_{i,l}$. Further define $T_{i,l}$ as the time spent on level $l$ or deeper: $T_{i,l} = \sum_{j=l}^{d} t_{i,j}$ with $d$ the depth of the pack. (Thus $T_i = T_{i,0}$.). We will assume a constant branching factor $b$ in the pack. Finally, we define $\bar{t}_l = \sum_i t_{i,l} / n$ with $n = b^d$. For simplicity, in the formulae we implicitly assume that $i$ always ranges from 1 to $n$ with $n$ the number of queries, unless explicitly specified otherwise. We then have

$$T_p = \max_i t_{i,0} + \sum_i T_{i,1} = \max_i t_{i,0} + \sum_{j=1}^{b}(\max_{i \in G_j} t_{i,1} + \sum_{i \in G_j} T_{i,2}) \qquad (4)$$

where $j = 1\ldots b$ is the index of a child of the root and $G_j$ is the set of indexes of the queries belonging to that child. Now define $K_0 = \max_i t_{i,0}/\bar{t}_0$ and define $K_1$ as the smallest number such that $\max_{i \in G_j} t_{i,1} \leq K_1 \bar{t}_{j,1}$ with $\bar{t}_{j,1} = \sum_{i \in G_j} t_{i,1}/b$. Note $1 \leq K_0, K_1 \leq b$. It then follows that

$$\sum_{j=1}^{b} \max_{i \in G_j} t_{i,1} \leq K_1 \sum_{j=1}^{b} \bar{t}_{j,1} = K_1 b \bar{t}_1 \qquad (5)$$

which allows us to rewrite Equation (4) into

$$T_p \leq K_0 \bar{t}_0 + K_1 b \bar{t}_1 + \sum_i T_{i,2} \qquad (6)$$

where the equality holds if $\max_{i \in G_j} t_{i,1}$ is equal in all $G_j$. The reasoning can be continued up till the lowest level of the pack, yielding

$$T_p \leq K_0 \bar{t}_0 + b K_1 \bar{t}_1 + b^2 K_2 \bar{t}_2 + \cdots + b^{d-1} K_{d-1} \bar{t}_{d-1} + \sum_i t_{i,d} \qquad (7)$$

and finally

$$T_p \leq K_0 \bar{t}_0 + b K_1 \bar{t}_1 + b^2 K_2 \bar{t}_2 + \cdots + b^{d-1} K_{d-1} \bar{t}_{d-1} + b^d \bar{t}_d \qquad (8)$$

with all $K_l$ between 1 and $b$. We will further simplify the comparison with $T_s$ by assuming $\forall l : K_l = 1$; the $K_l$ can then be dropped and the inequality becomes an equality (because all maxima must be equal):

$$T_p = \bar{t}_0 + b\bar{t}_1 + b^2 \bar{t}_2 + \cdots + b^{d-1}\bar{t}_{d-1} + b^d \bar{t}_d \qquad (9)$$

Note that for $T_s$ we have

$$T_s = b^d \bar{t}_0 + b^d \bar{t}_1 + b^d \bar{t}_2 + \cdots + b^d \bar{t}_{d-1} + b^d \bar{t}_d \qquad (10)$$

It is clear, then, that the speedup will be governed by how the $b^d \bar{t}_k$ terms compare to the $b^k \bar{t}_k$ terms. (In the worst case, where $K_k = b$, the latter become $b^{k+1}\bar{t}_k$.) We therefore introduce $R_{l,m}$ as follows:

$$R_{l,m} = \frac{\sum_{k=l}^{m} b^m \bar{t}_k}{\sum_{k=l}^{m} b^k \bar{t}_k} \qquad (11)$$

The $R$ coefficients are always between 1 (if $\bar{t}_m$ dominates) and $b^{m-l}$ (if $\bar{t}_l$ strongly dominates); for all $\bar{t}_l$ equal, $R_{l,m}$ is approximately $m - l$.

Further, similar to $c$ in our previous analysis, define

$$c_l = \frac{\sum_{k=0}^{l} b^k \bar{t}_k}{\sum_{k=l+1}^{d} b^k \bar{t}_k} \qquad (12)$$

Some algebra then gives
$$\frac{T_s}{T_p} = \frac{b^{d-l}c_l R_{0,l} + R_{l+1,d}}{c_l + 1} \tag{13}$$
which needs to hold for all $l$. We can interpret this as follows: for a certain level $l$, $c_l$ roughly reflects the speedup gained by the fact that the part up till level $l$ needs to be executed only once; the $R$ factors reflect the speedup obtained within these parts because of the pack mechanism.

This inequality holds for all $l$, hence we will find the best lower bound for the speedup factor by maximizing the right hand side. Note that $c_l$ increases and $b^{d-l}$ decreases monotonically with $l$. It is clear that if at some point $c_l$ becomes much larger than 1, a speedup factor of roughly $b^{d-l}$ is obtained. On the other hand, if $c_l$ is smaller than 1, then the behaviour of $b^{d-l}c_l$ is crucial. Now,

$$b^{d-l}c_l = \frac{\bar{t}_l + \frac{1}{b}\bar{t}_{l-1} + \cdots + \frac{1}{b^l}\bar{t}_0}{\bar{t}_d + \frac{1}{b}\bar{t}_{d-1} + \cdots + \frac{1}{b^{d-l-1}}\bar{t}_{l+1}}.$$

Our conclusion is similar to that for the one-level pack. If for some $l$, $c_l \gg 1$, i.e., if in the upper part of the pack (up till level $l$) computations take place that are so expensive that they dominate all computations below level $l$ (even taking into account that the latter are performed $b^{d-l}$ times more often), then a speedup of $b^{d-l}$ can be expected. If $c_l \ll 1$, which will usually be the case for all $l$ except those near $d$, the speedup can roughly be estimated as $\bar{t}_l/\bar{t}_d$. The maximum of all these factors will determine the actual speedup.

## 4. Adapting ILP Algorithms to Use Query Packs

In this section we discuss how the above execution method can be included in ILP algorithms, and illustrate this in more detail for two existing ILP algorithms. Experimental results concerning actual efficiency improvements this yields are presented in the next section.

### 4.1 Refinement of a Single Rule

Many systems for inductive logic programming use an algorithm that consists of repeatedly refining clauses. Any of these systems could in principle be rewritten to make use of a query pack evaluation mechanism and thus achieve a significant efficiency gain. We first show this for a concrete algorithm for decision tree induction, then discuss the more general case.

#### 4.1.1 Induction of Decision Trees

The first algorithm we discuss is TILDE (Blockeel & De Raedt, 1998), an algorithm that builds first-order decision trees. In a first-order decision tree, nodes contain literals that together with the conjunction of the literals in the nodes above this node (i.e., in a path from the root to this node) form the query that is to be run for an example to decide which subtree it should be sorted into. When building the tree, the literal (or conjunction of literals) to be put in one node is chosen as follows: given the query corresponding to a path from the root to this node, generate all refinements of this query (a refinement of a query is formed by adding one or more literals to the query); evaluate these refinements on the relevant subset of the data,[2] computing, e.g., the information gain (Quinlan, 1993a) yielded by the refinement; choose the best refinement; and put the literals that were added to the original clause to form this refinement in the node.

At this point it is clear that a lot of computational redundancy exists if each refinement is evaluated separately. Indeed all refinements contain exactly the same literals except those added during this single refinement step. Organising all refinements into one query pack, we obtain a query pack that essentially has only one level (the root immediately branches into leaves). When TILDE's lookahead facility is used (Blockeel & De Raedt, 1997), refinements form a lattice and the query pack may contain multiple (though usually few) levels.

Note that the root of these packs may consist of a conjunction of many literals, giving the pack a broom-like form. The more literals in the root of the pack, the greater the benefit of query pack execution is expected to be.

**Example 4** *Assume the node currently being refined has the following query associated with it:* ?- circle(A,C),leftof(A,C,D),above(A,D,E), *i.e., the node covers all examples A where there is a circle to the left of some other object which is itself above yet another object.*

*The query pack generated for this refinement could for instance be*

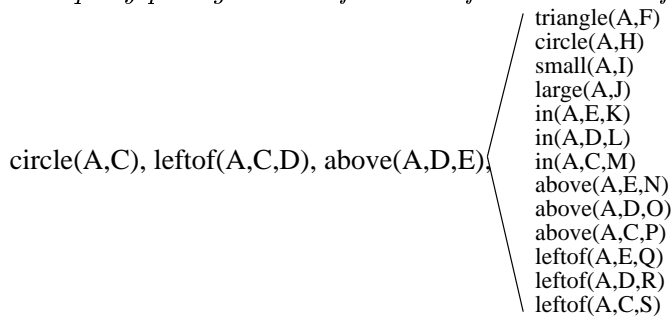

When evaluating this pack, all backtracking through the root of the pack (the "stick" of the broom) will happen only once, instead of once for each refinement. In other words: when evaluating queries one by one, for each query the Prolog engine needs to search once again for all objects C, D and E fulfilling the constraint circle(A,C), leftof(A,C,D), above(A,D,E); when executing a pack this search is done only once.

### 4.1.2 OTHER ALGORITHMS BASED ON RULE REFINEMENT

As mentioned, any ILP algorithm that consists of repeatedly refining clauses could in principle be rewritten to make use of a query pack evaluation mechanism and thus achieve a significant efficiency gain. Consider, e.g., a rule induction system performing an $A^*$ search through a refinement lattice, such as PROGOL (Muggleton, 1995). Since $A^*$ imposes a certain order in which clauses will be considered for refinement, it is hard to reorganise the computation at this level. However, when taking one node in the list of open nodes and producing all its refinements, the evaluation of the refinements involves executing all of them; this can be replaced by a pack execution, in which case a positive efficiency gain is guaranteed. In principle one could also perform several levels of refinement at this stage, adding all of the refinements to $A^*$'s queue; part of the efficiency of $A^*$ is then lost, but the pack execution mechanism is exploited to a larger extent. Which of these two effects is dominant will depend on the application: if most of the first-level refinements would be further refined anyway at some point during the search, clearly there will be a gain in executing a two-level pack; otherwise there may be a loss of efficiency. For instance, if executing a two-level pack takes $x$ times as much time as a one-level pack, it will bring an efficiency gain only if at least $x$ of the first level refinements would afterwards be refined themselves.

---

2. I.e., that subset of the original data set for which the parent query succeeded; or, in the decision tree context: the examples sorted into the node that is being refined.

### 4.2 Level-wise Frequent Pattern Discovery

An alternative family of data mining algorithms scans the refinement lattice in a breadth-first manner for queries whose frequency exceeds some user-defined threshold. The best-known instance of these level-wise algorithms is the APRIORI method for finding frequent item-sets (Agrawal et al., 1996). WARMR (Dehaspe & Toivonen, 1999) is an ILP variant of attribute-value based APRIORI.

Query packs in WARMR correspond to hash-trees of item-sets in APRIORI: both are used to store a subgraph of the total refinement lattice down to level $n$. The paths from the root down to level $n-1$ in that subgraph correspond to frequent patterns. The paths from root to the leaves at depth $n$ correspond to candidates whose frequency has to be computed. Like hash-trees in APRIORI, query packs in WARMR exploit massive similarity between candidates to make their evaluation more efficient. Essentially the WARMR algorithm starts with an empty query pack and iterates between pack evaluation and pack extension (see Figure 8). The latter is achieved by adding all potentially frequent refinements[3] of all leaves in the pack, i.e., adding another level of the total refinement lattice.

## 5. Experiments

The goal of this experimental evaluation is to empirically investigate the actual speedups that can be obtained by re-implementing ILP systems so that they use the pack execution mechanism. At this moment such re-implementations exist for the TILDE and WARMR systems, hence we have used these for our experiments. These re-implementations are available within the ACE data mining tool, available for academic use upon request.[4] We attempt to quantify (a) the speedup of packs w.r.t. to separate execution of queries (thus validating our complexity analysis), and (b) the total speedup that this can yield for an ILP system.

The data sets that we have used for our experiments are the following:

- The Mutagenesis data set : an ILP benchmark data set, introduced to the ILP community by Srinivasan et al. (1995), that consists of structural descriptions of 230 molecules that are to be classified as mutagenic or not. Next to the standard Mutagenesis data set, we also consider versions of it where each example occurs $n$ times;

---

3. Refinements found to be specialisations of infrequent queries cannot be frequent themselves, and are pruned consequently.
4. See http://www.cs.kuleuven.ac.be/~dtai/ACE/.

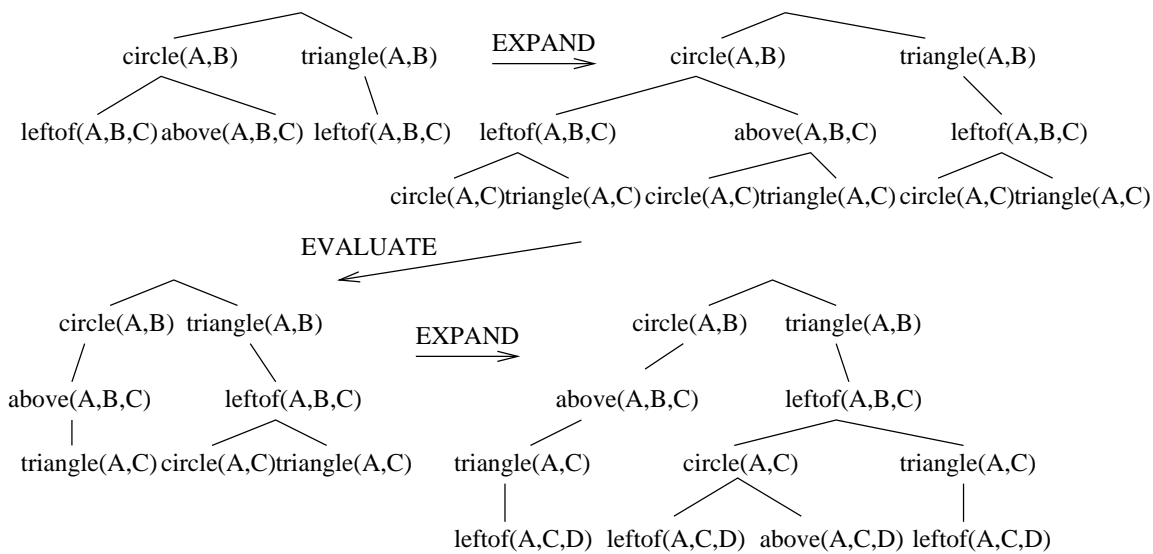

Figure 8: A sequence of 4 query packs in WARMR. Refinement of the above left query pack results in the 3-level pack above right. Removal of queries found infrequent during pack evaluation results in the bottom left pack. Finally, another level is added in a second query expansion step to produce the bottom right pack. This iteration between expansion and evaluation continues until the pack is empty.

this allows us to easily generate data sets of larger size where the average example and query complexity are constant and equal to those of the original data set.

- Bongard data sets : introduced in ILP by De Raedt and Van Laer (1995), the so-called "Bongard problems" are a simplified version of problems used by Bongard (1970) for research on pattern recognition. A number of drawings are shown containing each a number of elementary geometrical figures; the drawings have to be classified according to relations that hold on the figures in them. We use a Bongard problem generator to create data sets of varying size.

The experiments were run on SUN workstations: a Sparc Ultra-60 at 360 MHz for TILDE, a Sparc Ultra-10 at 333 Mhz for WARMR. TILDE and WARMR were run with their default settings, except where mentioned differently.

### 5.1 Tilde

We consider three different ways in which TILDE can be run in its ILPROLOG implementation:

1. No packs: the normal implementation of TILDE as described by Blockeel and De Raedt (1998), where queries are generated one by one and each is evaluated on all relevant examples. Since queries are represented as terms, each evaluation of a query involves a meta-call in Prolog.

2. Disjoint execution of packs: a query pack is executed in which all queries in the pack are put beside one another; i.e., common parts are not shared by the queries. The computational redundancy in executing such a pack is the same as that in executing all queries one after another; the main difference is that in this case all queries are compiled.

3. Packed execution of packs: a compiled query pack is executed where queries share as much as possible.

The most interesting information is obtained by comparing (a) the actual query evaluation time in settings 2 and 3: this gives a view of the efficiency gain obtained by the removal of redundant computation itself (we will abbreviate this as *exec* in the tables); and (b) the total execution time in settings 1 and 3: this provides an indication of how much is gained by implementing packs in an ILP system, taking all other effects into account (re-implementation of the computation of heuristics via a bit matrix, use of compiled queries instead of meta-calls, etc.), or in other words: what the net effect of the whole re-implementation is (indicated as *net* in the tables).

In a first experiment we used Bongard problems, varying (1) the size of the data sets; (2) the complexity of the target hypothesis; and (3) TILDE's `lookahead` parameter. The complexity of the target hypothesis can be small, medium, or none. In the latter case the examples are random, which causes TILDE to grow ever larger trees in an attempt to find a good hypothesis; the size of the final tree then typically depends on the size of the data set. The lookahead parameter is used to control the number of levels the pack contains; with lookahead $n$, packs of depth $n + 1$ are generated.

Table 1 gives an overview of results for the Bongard problems. The total induction time is reported, as well as (for pack-based execution mechanisms) the time needed for pack compilation and pack execution. Note that the total time includes not only pack compilation and execution, but also all other computations not directly related to packs (e.g., the computation of heuristics from the bitmatrix). The results can be interpreted as follows.

First of all, the table shows that significant speedups can be obtained by using the pack mechanism; net speedups of over a factor 5.5 are obtained, while the execution itself is up to 75 times faster compared to disjoint execution.

A further observation is that for more complex target hypotheses greater speedups are obtained. This can be explained by the broom-like form of the packs in TILDE. Complex target hypotheses correspond to deep trees, and refinement of a node at a lower level of such a tree yields a pack with a long clause before the branching, which in accordance with our previous analysis should yield a speedup closer to the branching factor $b$ in the case of lookahead 0 (and more generally, closer to $b^{l+1}$ for lookahead $l$, although the latter is much harder to achieve). Note that the maximum branching factor occurring in each pack is included in the table in column $bf$.

Finally, deeper packs also yield higher speedups, and this effect is larger for more complex theories. This is understandable considering the following. Let us call the clause that is being refined $c$. With lookahead $l$, conjunctions of $l + 1$ literals are added to the clause. In some cases the first of these $l + 1$ literals may fail immediately, which causes this branch of the pack to have almost no execution time, while cutting away $b^l$ queries. Remember that according to our analysis, the speedup can in the limit approximate $b^l$ if the complexity of clause $c$ dominates over the complexity of the rest of the pack; such "early failing branches" in the pack cause the actual situation to approximate closer this ideal case.

| LA | bf | original | disjoint | | | packed | | | speedup | |
|---|---|---|---|---|---|---|---|---|---|---|
| | | | total | comp | exec | total | comp | exec | net | exec |
| | | | | Simple target hypothesis | | | | | | | |
| | | | | 1007 examples | | | | | | | |
| 0 | 16 | *0.74* | *0.62* | 0.14 | **0.13** | *0.49* | 0.05 | **0.07** | *1.51* | **1.86** |
| 1 | 24 | *2.44* | *1.64* | 0.35 | **0.45** | *1.09* | 0.14 | **0.11** | *2.24* | **4.09** |
| 2 | 18 | *7.49* | *4.07* | 0.8 | **1.57** | *2.15* | 0.27 | **0.16** | *3.48* | **9.81** |
| 3 | 21 | *29.9* | *16.52* | 3.65 | **7.26** | *7.18* | 1.26 | **0.28** | *4.17* | **25.9** |
| | | | | 2473 examples | | | | | | | |
| 0 | 16 | *1.82* | *1.43* | 0.17 | **0.34** | *1.13* | 0.07 | **0.16** | *1.61* | **2.13** |
| 1 | 24 | *5.72* | *3.34* | 0.34 | **1.17** | *2.24* | 0.11 | **0.3** | *2.55* | **3.9** |
| 2 | 18 | *17.2* | *8.45* | 0.78 | **3.95** | *4.4* | 0.27 | **0.39** | *3.92* | **10.1** |
| 3 | 21 | *69.8* | *33.0* | 3.57 | **17.5** | *13.7* | 1.13 | **0.69** | *5.11* | **25.4** |
| | | | | 4981 examples | | | | | | | |
| 0 | 19 | *3.69* | *2.72* | 0.29 | **0.67** | *2.16* | 0.12 | **0.32** | *1.71* | **2.09** |
| 1 | 24 | *11.4* | *6.22* | 0.35 | **2.41** | *4.17* | 0.13 | **0.63** | *2.74* | **3.83** |
| 2 | 18 | *34.7* | *16.0* | 0.74 | **8.14** | *8.24* | 0.25 | **0.88** | *4.21* | **9.25** |
| 3 | 21 | *142* | *62.4* | 3.61 | **36.5** | *24.9* | 1.09 | **1.45** | *5.69* | **25.1** |
| | | | | Medium complexity target hypothesis | | | | | | | |
| | | | | 1031 examples | | | | | | | |
| 0 | 19 | *1.01* | *0.93* | 0.29 | **0.18** | *0.66* | 0.11 | **0.07** | *1.53* | **2.57** |
| 1 | 21 | *3.26* | *2.8* | 0.98 | **0.56** | *1.66* | 0.35 | **0.14** | *1.96* | **4** |
| 2 | 15 | *6.36* | *3.47* | 0.68 | **1.22** | *1.95* | 0.25 | **0.15** | *3.26* | **8.13** |
| 3 | 18 | *27.2* | *14.6* | 3.75 | **5.75** | *6.71* | 1.20 | **0.27** | *4.06* | **21.3** |
| | | | | 2520 examples | | | | | | | |
| 0 | 22 | *3.16* | *2.82* | 0.89 | **0.62** | *1.91* | 0.3 | **0.24** | *1.65* | **2.58** |
| 1 | 24 | *8.38* | *5.88* | 1.5 | **1.86** | *3.3* | 0.44 | **0.41** | *2.54* | **4.54** |
| 2 | 27 | *38.5* | *29.8* | 13.14 | **9.52** | *10.3* | 2.44 | **0.6** | *3.73* | **15.9** |
| 3 | 18 | *124* | *58.02* | 10.3 | **28.6** | *23.9* | 3.00 | **1.11** | *5.21* | **25.7** |
| | | | | 5058 examples | | | | | | | |
| 0 | 25 | *6.35* | *5.41* | 1.47 | **1.3** | *3.73* | 0.56 | **0.53** | *1.70* | **2.45** |
| 1 | 24 | *18.14* | *12.98* | 3.2 | **4.15** | *7.5* | 0.93 | **0.91** | *2.42* | **4.56** |
| 2 | 27 | *119* | *93.2* | 38.1 | **31.0** | *35.3* | 9.09 | **1.7** | *3.36* | **18.2** |
| 3 | 27 | *384* | *275* | 108 | **89.1** | *106* | 25.9 | **2.83** | *3.62* | **31.5** |
| | | | | No target hypothesis | | | | | | | |
| | | | | 1194 examples | | | | | | | |
| 0 | 28 | *4.74* | *6.65* | 3.34 | **0.94** | *3.93* | 0.98 | **0.20** | *1.21* | **4.70** |
| 1 | 24 | *16.32* | *21.29* | 10.97 | **2.24** | *11.65* | 3.41 | **0.31** | *1.40* | **7.23** |
| 2 | 24 | *87.5* | *130* | 82.3 | **13.8** | *54.7* | 20.4 | **0.57** | *1.60* | **24.1** |
| 3 | 30 | *373* | *519* | 316 | **61.1** | *220* | 74.9 | **1.34** | *1.70* | **45.6** |
| | | | | 2986 examples | | | | | | | |
| 0 | 31 | *12.7* | *16.5* | 7.04 | **2.68** | *9.8* | 2.16 | **0.56** | *1.30* | **4.79** |
| 1 | 36 | *65.1* | *83.7* | 42.9 | **10.7** | *42.47* | 11.2 | **1.14** | *1.53* | **9.39** |
| 2 | 33 | *430* | *606* | 396 | **84** | *211.3* | 82.58 | **2.57** | *2.03* | **32.6** |
| 3 | 33 | *1934* | *2592* | 1610 | **375** | *946* | 332 | **6.58** | *2.04* | **57.0** |
| | | | | 6013 examples | | | | | | | |
| 0 | 31 | *25.3* | *30.3* | 11.8 | **5.53** | *18.3* | 3.53 | **1.27** | *1.38* | **4.35** |
| 1 | 39 | *154* | *198* | 91.2 | **33.4** | *99.9* | 22.0 | **3.13** | *1.54* | **10.7** |
| 2 | 39 | *1185* | *1733* | 1076 | **358** | *504* | 197 | **9** | *2.35* | **39.8** |
| 3 | 42 | *4256* | *6932* | 4441 | **1091** | *2006* | 695 | **14.5** | *2.12* | **75.4** |

Table 1: Timings of TILDE runs on the Bongard data sets. $LA$ = lookahead setting; $bf$ = maximum branching factor. Reported times (in seconds) are the total time needed to build a tree, and the time spent on compilation respectively execution of packs.

| LA | original | disjoint | | | packed | | | speedup ratio | |
|---|---|---|---|---|---|---|---|---|---|
| | | total | comp | exec | total | comp | exec | *net* | **exec** |
| Regression, 230 examples | | | | | | | | | |
| 0 | *31.5* | *52.9* | 1.96 | **25.5** | *45.5* | 1.02 | **19.25** | *0.69* | **1.33** |
| 1 | *194.99* | *248* | 55.9 | **109** | *107* | 12.6 | **16.6** | *1.82* | **6.53** |
| 2 | *2193* | – | – | – | *891* | 192 | **32.0** | *2.46* | – |
| Classification, 230 examples | | | | | | | | | |
| 0 | *27.6* | *27.3* | 1.83 | **4.71** | *25.4* | 1.13 | **3.42** | *1.09* | **1.38** |
| 1 | *38.02* | *40.3* | 7.55 | **9.09** | *30.6* | 3.11 | **3.65** | *1.24* | **2.49** |
| 2 | *638* | – | – | – | *149* | 74.3 | **6.16** | *4.2* | – |

Table 2: Timings of TILDE runs for Mutagenesis. A − in the table indicates that that run ended prematurely.

We have also run experiments on the Mutagenesis data set (Table 2), both in a regression and a classification setting. Here, query packs are much larger than for the Bongard data set (there is a higher branching factor); with a lookahead of 2 the largest packs had over 20000 queries. For these large packs a significant amount of time is spent compiling the pack, but even then clear net speedups are obtained.[5] A comparison of execution times turned out infeasible because in the disjoint execution setting the pack structures consumed too much memory.

### 5.2 Warmr

#### 5.2.1 USED IMPLEMENTATIONS

For WARMR we consider the following implementations:

1. No packs: the normal implementation of WARMR, where queries are generated, and for all examples the queries are evaluated one by one.

2. With packs: An implementation where first all queries for one level are generated and put into a pack, and then this pack is evaluated on each example.

#### 5.2.2 DATASETS

**Mutagenesis** We used the Mutagenesis dataset of 230 molecules, where each example is repeated 10 times to make more accurate timings possible and to have a better idea of the effect on larger datasets. We used three different language biases. 'small' is a language bias that was chosen so as to generate a limited number of refinements (i.e., a relatively small branching factor in the search lattice); this allows us to generate query packs that are relatively deep but narrow. 'medium' and 'large' use broader but more shallow packs. Table 3 summarises the number of queries and the number of frequent queries found for each level in the different languages.

---

5. In one case, with a relatively small pack, the system became slower. The timings indicate that this is not due to the compilation time, but to other changes in the implementation which for this relatively simple problem were not compensated by the faster execution of the packs.

|  | Mutagenesis | | | | | |
|---|---|---|---|---|---|---|
| Level | `small` | | `medium` | | `large` | |
|  | Queries | Frequent | Queries | Frequent | Queries | Frequent |
| 1 | 8 | 5 | 37 | 26 | 45 | 31 |
| 2 | 60 | 14 | 481 | 48 | 1071 | 211 |
| 3 | 86 | 24 | 688 | 114 | 3874 | 1586 |
| 4 | 132 | 31 | 699 | 253 | | |
| 5 | 37 | 21 | 697 | 533 | | |
| 6 | 29 | 18 | 1534 | 1149 | | |
| 7 | 23 | 15 | – | – | | |
| 8 | 17 | 12 | – | – | | |
| 9 | 4 | 4 | – | – | | |

Table 3: Number of queries for the Mutagenesis experiment with WARMR.

**Bongard** We use Bongard-6013 for experiments with WARMR as this system does not construct a theory and hence the existence of a simple theory is not expected to make much difference.

### 5.2.3 RESULTS

In Tables 4, 5 and 6 the execution times of WARMR on Mutagenesis are given, with maximal search depth varying from 3 for the large language to 9 levels for the small language. Here, 'total' is the total execution time and 'exec' is the time needed to test the queries against the examples. In Table 7 the execution times of WARMR on Bongard are given.

### 5.2.4 DISCUSSION

The execution time of WARMR has a large component that is not used to evaluate queries. This is caused by the fact that WARMR needs to do a lot of administrative work. In particular, theta-subsumption tests should be done on the queries to check wether a query is equivalent to another candidate, or if a query is a specialisation of an infrequent one. In the propositional case (the APRIORI algorithm), these tests are very simple, but in the first order case they require exponential time in the size of the queries. Of course, when using larger datasets, the relative contribution of these administrative costs will decrease proportionally. It can be observed that at deeper levels, these costs are less for the setting using packs. One of the causes is the fact that the no-packs version also uses more memory than the packs setting (and hence causes proportionally more memory management).

Here again, the most important numbers are the speedup factors for the execution of queries. Speedup factors of query execution do not always increase with increasing depth of the packs, in contrast to Tilde where larger packs yielded higher speedups. At first sight we found this surprising; however it becomes less so when the following observation is made. When refining a pack into a new pack by adding a level, Warmr prunes away branches that lead only to infrequent queries. There are thus two effects when adding a level to a pack: one is the widening of the pack at the lowest level (at least on the first few levels, a new pack typically has more leaves than the previous one), the second is the narrowing of the pack as a whole (because of pruning). Since the speedup obtained by using packs largely depends on the branching factor of the pack, speedup factors can be expected to decrease when the narrowing effect is stronger than the widening-at-the-bottom effect. This can be seen, e.g, in the small-mutagenesis experiment, where at the deepest levels queries are becoming less frequent. For the mutagenesis experiment with the medium size language, query execution speedup factors are larger as the number of queries increases much faster. For the mutagenesis experiment with the large language, it is the total speedup that is large, as the language generates so many queries that the most time-consuming part becomes the administration and storage in memory. The packs version is much faster as it stores the queries in trees, requiring significantly less memory.

| Level | No packs total | exec | With packs ILPROLOG total | exec | speedup ratio net | exec |
|---|---|---|---|---|---|---|
| 1 | 0.35 | **0.23** | 0.18 | **0.15** | *1.94* | **1.53** |
| 2 | 6.27 | **5.60** | 4.56 | **4.12** | *1.38* | **1.36** |
| 3 | 36.93 | **31.49** | 14.01 | **9.87** | *2.64* | **3.19** |
| 4 | 117.33 | **84.45** | 45.14 | **16.27** | *2.60* | **5.19** |
| 5 | 215.95 | **104.36** | 129.37 | **20.78** | *1.67* | **5.02** |
| 6 | 336.35 | **111.28** | 249.41 | **22.39** | *1.35* | **4.97** |
| 7 | 569.14 | **115.80** | 497.86 | **24.63** | *1.14* | **4.70** |
| 8 | 902.72 | **120.99** | 831.30 | **25.98** | *1.09* | **4.66** |
| 9 | 1268.16 | **119.60** | 1148.23 | **32.28** | *1.10* | **3.71** |

Table 4: Results for WARMR on the Mutagenesis dataset using a small language.

| Level | No packs total | exec | With packs ILPROLOG total | exec | speedup ratio net | exec |
|---|---|---|---|---|---|---|
| 1 | 2.58 | **2.27** | 2.16 | **2.09** | *1.19* | **1.09** |
| 2 | 112.98 | **42.32** | 34.35 | **13.39** | *3.29* | **3.16** |
| 3 | 735.19 | **128.67** | 262.83 | **34.70** | *2.80* | **3.71** |
| 4 | 4162.15 | **287.72** | 1476.06 | **54.10** | *2.82* | **5.32** |
| 5 | 17476.98 | **444.44** | 6870.16 | **73.11** | *2.54* | **6.08** |
| 6 | 65138.72 | **866.85** | 25921.73 | **104.81** | *2.51* | **8.27** |

Table 5: Results for WARMR on the Mutagenesis dataset using a medium language.

| Level | No packs total | exec | With packs ILPROLOG total | exec | speedup ratio net | exec |
|---|---|---|---|---|---|---|
| 1 | 2.82 | **2.42** | 2.28 | **2.11** | *1.24* | **1.15** |
| 2 | 408.85 | **102.38** | 102.29 | **50.67** | *4.00* | **2.02** |
| 3 | 27054.33 | **1417.76** | 3380.19 | **370.44** | *8.00* | **3.83** |

Table 6: Results for WARMR on the Mutagenesis dataset using a large language.

| Level | No packs | | With packs ilProlog | | speedup ratio | |
|---|---|---|---|---|---|---|
| | total | exec | total | exec | *net* | **exec** |
| 1 | 0.24 | **0.22** | 0.24 | **0.23** | *1.00* | **0.96** |
| 2 | 0.83 | **0.75** | 0.77 | **0.68** | *1.08* | **1.10** |
| 3 | 3.28 | **2.82** | 2.34 | **1.92** | *1.40* | **1.47** |
| 4 | 11.56 | **9.31** | 6.08 | **4.28** | *1.90* | **2.18** |
| 5 | 38.34 | **28.11** | 16.20 | **8.15** | *2.37* | **3.45** |
| 6 | 75.51 | **46.97** | 36.57 | **12.22** | *2.06* | **3.84** |
| 7 | 135.64 | **71.60** | 68.96 | **15.59** | *1.97* | **4.59** |
| 8 | 186.23 | **84.93** | 102.46 | **17.82** | *1.82* | **4.77** |
| 9 | 210.82 | **88.97** | 120.76 | **18.52** | *1.75* | **4.80** |
| 10 | 216.61 | **89.38** | 125.84 | **18.88** | *1.72* | **4.73** |

Table 7: Warmr results on Bongard.

### 5.3 Comparison with Other Engines

Implementing a new special-purpose Prolog engine, different from the already existing ones, carries a risk: given the level of sophistication of popular Prolog engines, it is useful to check whether the new engine performs comparably with these existing engines, at least for the tasks under consideration here. The efficiency gain obtained through query pack execution should not be offset by a less efficient implementation of the engine itself.

Originally the Tilde and Warmr systems were implemented in MasterProLog. In an attempt to allow them to run on other platforms, parts of these systems were re-implemented into a kind of "generic" Prolog from which implementations for specific Prolog engines (SICStus, ilProlog) can easily be derived (the low level of standardisation of Prolog made this necessary). Given this situation, there are two questions to be answered:

| Data set     | LA | MasterProLog | ilProlog(original) | ilProlog(packs) |
|--------------|----|--------------|---------------------|------------------|
| Bongard-1194 | 0  | 7.8          | 4.74                | 3.93             |
| Bongard-2986 | 0  | 17.8         | 12.7                | 9.8              |
| Bongard-6013 | 0  | 35           | 25                  | 18               |
| Bongard-1007 | 0  | 0.77         | 0.74                | 0.49             |
| Bongard-2473 | 0  | 2.07         | 1.82                | 1.13             |
| Bongard-4981 | 0  | 4.1          | 3.7                 | 2.2              |
| Bongard-1007 | 2  | 7.1          | 7.5                 | 2.2              |
| Bongard-2473 | 2  | 17.7         | 17.2                | 4.4              |
| Bongard-4981 | 2  | 38           | 35                  | 8.2              |

Table 8: ilProlog compared to other engines (times in seconds) for several data sets and lookahead settings.

(a) does the move from MasterProLog to other Prolog engines influence performance in a negative way; and (b) does the performance loss, if any, reduce the performance improvements due to the use of packs?

Tilde and Warmr have been tuned for fast execution on MasterProLog and ilProlog but not for SICStus, which makes a comparison with the latter unfair; therefore we just report on the former 2 engines. Table 8 shows some results. These confirm that ilProlog is competitive with state-of-the-art Prolog engines.

### 5.4 Summary of Experimental Results

Our experiments confirm that (a) query pack execution in itself is much more efficient than executing many highly similar queries separately; (b) existing ILP systems (we use Tilde and Warmr as examples) can use this mechanism to their advantage, achieving significant speedups; and c) although a new Prolog engine is needed to achieve this, the current state of development of this engine is such that with respect to execution speed it can compete with state-of-the-art engines. Further, the experiments are consistent with our complexity analysis of the execution time of packs.

## 6. Related Work

The re-implementation of Tilde is related to the work by Mehta et al. (1996) who were the first to describe the "examples in outer loop" strategy for decision tree induction. The query pack execution mechanism, here described from the Prolog execution point of view, can also be seen as a first-order counterpart of Apriori's mechanism for counting item-sets (Agrawal et al., 1996).

Other lines of work on efficiency improvements for ILP involves stochastic methods which trade a certain amount of optimality for efficiency by, e.g., evaluating clauses on a sample of the data set instead of the full data set (Srinivasan, 1999), exploring the clause search space in a random fashion (Srinivasan, 2000), or stochastically testing whether a query succeeds on an example (Sebag & Rouveirol, 1997). The first of these is entirely orthogonal to query pack execution and can easily be combined with it.

The idea of optimising sets of queries instead of individual queries has existed for a while in the database community. The typical context considered in earlier research on multi-query optimisation (e.g., Sellis, 1988) was that of a database system that needs to handle disjunctions of conjunctive queries, or of a server that may receive many queries from different clients in a brief time interval. If several of these queries are expected to compute the same intermediary relations, it may be more efficient to materialise these relations instead of having them recomputed for each query. Data mining provides in a sense a new context for multi-query optimisation, in which the multi-query optimisation approach is at the same time easier (the similarities among the queries are more systematic, so one need not look for them) and more promising (given the huge number of queries that may be generated at once).

Tsur et al. (1998) describe an algorithm for efficient execution of so-called query *flocks* in this context. Like our query pack execution mechanism, the query flock execution mechanism is inspired to some extent by APRIORI and is set in a deductive database setting. The main difference between our query packs and the query flocks described by Tsur et al. (1998) is that query packs are more hierarchically structured and the queries in a pack are structurally less similar than the queries in a flock. (A flock is represented by a single query with placeholders for constants, and is equal to the set of all queries that can be obtained by instantiating the placeholders to constants. Flocks could not be used for the applications we consider here.)

Dekeyser and Paredaens (2001) describe work on multi-query optimisation in the context of relational databases. They also consider tree-like structures in which multiple queries are combined; the main difference is that their trees are rooted in one single table from which the queries select tuples, whereas our queries correspond to joins of multiple tables. Further, Dekeyser and Paredaens define a cost measure for trees as well as operators that map trees onto semantically equivalent (but less costly) trees, whereas we have considered only the creation of packs and an efficient top-down execution mechanism for them. Combining both approaches seems an interesting topic for further research.

Finally, other optimisation techniques for ILP have been proposed that exploit results from program analysis (Santos Costa et al., 2000; Blockeel et al., 2000) or from propositional data mining technology (Blockeel et al., 1999). These are complementary to our pack execution optimisation. Especially the approach of Blockeel et al. (1999) can easily be combined with our pack mechanism. The techniques discussed by Santos Costa et al. (2000) and Blockeel et al. (2000) involve optimisations for single query execution, some of which can to some extent be upgraded to the pack setting. This is future work.

## 7. Conclusions

There is a lot of redundancy in the computations performed by most ILP systems. In this paper we have identified a source of redundancy and proposed a method for avoiding it: execution of query packs. We have discussed how query pack execution can be incorporated in ILP systems. The query pack execution mechanism has been implemented in a new Prolog system called ILPROLOG and dedicated to data mining tasks, and two ILP systems have been re-implemented to make use of the mechanism. We have experimentally evaluated these re-implementations, and the results of these experiments confirm that large speedups may be obtained in this way. We conjecture that the query pack execution mechanism can be incorporated in other ILP systems and that similar speedups can be expected.

The problem setting in which query pack execution was introduced is very general, and allows the technique to be used for any kind of task where many queries are to be executed on the same data, as long as the queries can be organised in a hierarchy.

Future work includes further improvements to the ILPROLOG engine and the implementation of techniques that will increase the suitability of the engine to handle large data sets. In the best case one might hope to combine techniques known from database optimisation and program analysis with our pack execution mechanism to further improve the speed of ILP systems.

## Acknowledgements

Hendrik Blockeel is a post-doctoral fellow of the Fund for Scientific Research (FWO) of Flanders. Jan Ramon is funded by the Flemish Institute for the Promotion of Scientific Research in Industry (IWT). Henk Vandecasteele was funded in part by the FWO project G.0246.99, "Query languages for database mining". The authors thank Luc De Raedt for his influence on this work, Ashwin Srinivasan for suggesting the term "query packs", the anonymous reviewers for their useful comments, and Kurt Driessens for proofreading this text. This work was motivated in part by the Esprit project 28623, Aladin.

## Appendix A. Preparing the Query for the Meta-interpreter

Note that the following preprocessor assumes that the pack of the form `a, (b, (c or d or e) or f or g, (h or i or j))` was already transformed to the form `a , or([(b, or([c,d,e])), f, (g, or([h,i,j]))])`.

```
preprocess((A,B),(A,NewB),PrevNode,NodeNr0,LeafNr0,BranchNr,NodeNr1,LeafNr1):- !,
       preprocess(B,NewB,PrevNode,NodeNr0,LeafNr0,BranchNr,NodeNr1,LeafNr1).
preprocess(or(Querys),or(NQuerys,PrevNode,NodeNr0,BranchNr,Length),
                    PrevNode,NodeNr0,LeafNr0,BranchNr, NodeNr1,LeafNr1):- !,
       NodeNr2 is NodeNr0 + 1,
       preprocessbranches(Querys,NQuerys,NodeNr0,NodeNr2,LeafNr0,
                    1,NodeNr1,LeafNr1,Length).
preprocess(A,(A,leaf(PrevNode,BranchNr,LeafNr0)),
                    PrevNode,NodeNr0,LeafNr0, BranchNr,NodeNr0,LeafNr1):-
       LeafNr1 is LeafNr0 + 1.

preprocessbranches([],[],_,NodeNr,LeafNr,BranchNr, NodeNr,LeafNr,BranchNr).
preprocessbranches([Query|Querys],[NewQuery|NewQuerys],PrevNode,
                    NodeNr0,LeafNr0,BranchNr, NodeNr1,LeafNr1,Length):-
       preprocess(Query,NewQuery,
                    PrevNode,NodeNr0,LeafNr0,BranchNr, NodeNr2,LeafNr2),
       BranchNr1 is BranchNr + 1,
       preprocessbranches(Querys,NewQuerys, PrevNode,
                    NodeNr2,LeafNr2,BranchNr1, NodeNr1,LeafNr1,Length).
```